\pdfoutput=1

\documentclass[11pt]{article}

\usepackage[]{acl}

\usepackage{times}
\usepackage{latexsym}

\usepackage{amssymb}
\usepackage{amsmath}
\usepackage{multirow}
\usepackage{booktabs}
\usepackage{graphicx}
\usepackage{soul}

\usepackage{algorithm} 
\usepackage{algpseudocode} 
\makeatletter
\renewcommand{\ALG@beginalgorithmic}{\small}
\makeatother
\algnewcommand\algorithmicinput{\textbf{Input:}}
\algnewcommand\algorithmicoutput{\textbf{Output:}}
\algnewcommand\INPUT{\item[\algorithmicinput]}
\algnewcommand\OUTPUT{\item[\algorithmicoutput]}

\usepackage{tikz}
\usepackage{pgfplots}
\usetikzlibrary{positioning,matrix,calc,arrows.meta,decorations.pathreplacing,shapes.geometric}
\definecolor{mygreen}{RGB}{46,139,87}
\definecolor{myred}{RGB}{255,152,150}
\definecolor{myblue}{RGB}{30,144,255}
\definecolor{myyellow}{RGB}{219,219,141}
\definecolor{mybrown}{RGB}{197,157,148}
\usepackage{pgfplotstable}

\usepackage{bm}
\usepackage{url}
\usepackage{booktabs}
\usepackage{subcaption}

\usepackage[T1]{fontenc}

\usepackage[utf8]{inputenc}

\usepackage{microtype}

\newcommand\LegendImage[1]{
	\draw[%
	/pgfplots/mesh=false,%
	bar width=3pt,%
	bar shift=0pt,%
	mark repeat=2,%
	mark phase=2,#1]
	plot coordinates {
		(0cm,0cm)
		(0.3cm,0cm)
		(0.6cm,0cm)%
	};
}
\newcommand\LegendEntry[1]{\node[anchor=west,black,font=\footnotesize,inner xsep=2pt]{#1};}
\newcommand{\confidenceband}[5][]{ 
\pgfplotstableread{#2}\datatable
    \addplot [draw=none, stack plots=y, forget plot] table [
        x={#3},
        y expr=\thisrow{#5}
    ] {\datatable};

    \addplot [draw=none, fill=gray!40, stack plots=y, area legend, #1] table [
        x={#3},
				y expr=\thisrow{#4}-\thisrow{#5}
    ] {\datatable} \closedcycle;

    \addplot [forget plot, stack plots=y,draw=none] table [x={#3}, y expr=-\thisrow{#4}] {\datatable};
}

\title{Hybrid Alignment Training for Large Language Models}

\author{Chenglong Wang\textsuperscript{\rm 1},
    Hang Zhou\textsuperscript{\rm 1},
    Kaiyan Chang\textsuperscript{\rm 1},
    Bei Li\textsuperscript{\rm 1},
    Yongyu Mu\textsuperscript{\rm 1}, \\
    \bf{Tong Xiao}\textsuperscript{\rm 1,3}\footnotemark[1], 
    \bf{Tongran Liu}\textsuperscript{\rm 2}, and
    \bf{Jingbo Zhu}\textsuperscript{\rm 1,3} \\
    \textsuperscript{\rm 1} School of Computer Science and Engineering, Northeastern University, Shenyang, China \\
    \textsuperscript{\rm 2} CAS Key Laboratory of Behavioral Science, Institute of Psychology, CAS, Beijing, China \\
    \textsuperscript{\rm 3} NiuTrans Research, Shenyang, China \\
    \ttfamily{\{clwang1119, ctrl.hang\}@gmail.com},\\
    \ttfamily{\{xiaotong, zhujingbo\}@mail.neu.edu.cn}
}

\begin{document}
\maketitle
\begin{abstract}
Alignment training is crucial for enabling large language models (LLMs) to cater to human intentions and preferences.
It is typically performed based on two stages with different objectives: instruction-following alignment and human-preference alignment.
However, aligning LLMs with these objectives in sequence suffers from an inherent problem: the objectives may conflict, and the LLMs cannot guarantee to simultaneously align with the instructions and human preferences well.
To response to these, in this work, we propose a \textbf{H}y\textbf{b}rid \textbf{A}lignment \textbf{T}raining (\textsc{Hbat}) approach, based on alternating alignment and modified elastic weight consolidation methods.
The basic idea is to alternate between different objectives during alignment training, so that better collaboration can be achieved between the two alignment tasks. 
We experiment with \textsc{Hbat} on summarization and dialogue tasks.
Experimental results show that the proposed \textsc{Hbat} can significantly outperform all baselines.
Notably, \textsc{Hbat} yields consistent performance gains over the traditional two-stage alignment training when using both proximal policy optimization and direct preference optimization.

\footnotetext[1]{Corresponding author.}

\end{abstract}

\section{Introduction}
Alignment training is a key technique to ensure that the behaviors of large language models (LLMs) are consistent with human intentions and preferences \cite{ouyang2022training, wang2023aligning}.
It typically involves two stages: 1) using human-labeled data to train pre-trained LLMs via a supervised training method, which enables LLMs to understand human intentions and follow the instructions (call it \textit{instruction-following alignment}), and 2) employing approaches like proximal policy optimization (PPO) \cite{schulman2017proximal} and direct preference optimization (DPO) \cite{rafailov2023direct} to learn preferences from human feedbacks (call it \textit{human-preference alignment}). 
This paradigm has achieved promising results on several downstream tasks, such as dialogue \cite{chatgpt2022, dubois2023alpacafarm, wang2023esrl}, summarization \cite{stiennon2020learning, lee2023rlaif}, and machine translation \cite{ramos2023aligning}.

However, this two-stage alignment training has its inherited limitation: the optimization objectives are different for each stage, which can make an optimization conflict \cite{french1999catastrophic, liu2021conflict}.
Such limitation could result in an inferiorly aligned LLM in real-world scenarios. 
This phenomenon is also described in \citet{ouyang2022training}'s work, which is referred to as alignment tax.


To mitigate this limitation, in this work, we propose a \textbf{H}y\textbf{b}rid \textbf{A}lignment \textbf{T}raining (\textsc{Hbat}) approach, which offers a refinement of the collaboration among instruction-following alignment and human-preference alignment by using the following two methods.
For one, inspired by interactive methods in multi-objective optimization \cite{miettinen2008introduction, xin2018interactive}, we propose an alternating alignment method, where the human-preference alignment acts as a decision maker and continuously interacts with the instruction-following alignment to achieve a preferred alignment.
Specifically, we divide the instruction-following and human-preference training set into equal portions of mutually exclusive subsets, respectively.
Then, we rearrange these subsets in alternating orders during alignment training.
Furthermore, we introduce a modified Elastic Weight Consolidation (EWC) \cite{kirkpatrick2017overcoming} to alternating alignment.
EWC is a method to dynamically impose an appropriate constraint on each parameter when training a model with a new optimization objective, thereby easing an optimization conflict with the previous objective.

We experiment with the proposed \textsc{Hbat} on summarization and dialogue tasks based on LLaMA2-7B and LLaMA2-13B models \cite{touvron2023llama}.
Experimental results show that \textsc{Hbat} can significantly surpass all baselines.
Notably, based on the LLaMA2-13B model, \textsc{Hbat} can yield a +2.26 ROUGE-L points improvement for the summarization task, compared to the traditional RLHF.
Additionally, our \textsc{Hbat} significantly outperforms the SFT over 21.01 GPT-4 win rate points on the dialogue task based on the LLaMA2-13B model.
Furthermore, \textsc{Hbat} is orthogonal to other optimized alignment approaches.
For instance, when armed with ESRL \cite{wang2023esrl}, our \textsc{Hbat} gains an additional improvement of 2.59 GPT-4 win rate points on the summarization task.

\section{Related Work}
\paragraph{Alignment Training for LLMs.}
Recently, many efforts have been made to improve the LLM alignment for different tasks \cite{stiennon2020learning, nakano2021webgpt, wang2023making, hu2023aligning}.
These works mainly focused on optimizing each stage of alignment training, including instruction-following alignment (also referred to as SFT) and human-preference alignment (also referred to as RLHF). 
For example, \citet{zhou2023lima} designed data selection schemes to provide high-quality instruction-following data.
Moreover, \citet{wang2022self} proposed an efficient approach for producing instruction-following data.
Likewise, some works aimed to efficiently produce human-preference data \cite{lee2023rlaif, dubois2023alpacafarm, wang2023learning}.
Apart from the training data improvements, another line of improving the alignment training is to explore better reward models and optimization objectives, such as the use of fine-grained reward models \cite{coste2023reward, wu2023fine}, the integration of a prior knowledge in training reward models \cite{zhou2024prior}, and the design of direct preference optimization objective \cite{rafailov2023direct}.
Although previous works improve the performance of instruction-following alignment and human-preference alignment, they rarely consider the optimization conflict limitation between them.
Researchers have been aware of this \cite{ouyang2022training}, but it is still rare to see studies on this issue.

\paragraph{Multi-objective Optimization.}
Multi-objective optimization problem involves optimizing multiple optimization objectives simultaneously \cite{hwang2012multiple}. 
However, there does not typically exist a feasible solution that minimizes all loss functions.
Therefore, researchers always explored a Pareto optimal solution that cannot be improved in any of the objectives without impairing at least one of the other objectives.
Recent works on this exploration could be classified into three groups.
The first group focused on Pareto dominance-based method.
This method maintains the individual elements of the solution vectors as independent during optimization \cite{cheng2015many, wu2015multi}.
The second group tended to design an quality indicator, such as hypervolume \cite{bader2011hype} and R2 \cite{wagner2013preference}, to act as a proxy objective instead of optimization objectives.
The third group that has attracted attention commonly aimed to solve multi-objective optimization problems through an interactive method.
A typical interactive method requires a decision maker to offer preference information, which allows to search for the most preferred Pareto optimal solution after each optimization \cite{xin2018interactive, misitano2021desdeo, pereira2022review}.

Although the alignment training is not a standard multi-objective optimization problem, its goal remains consistent, \textit{i.e.,} seeking an aligned LLM that simultaneously aligns instructions and human preferences well.

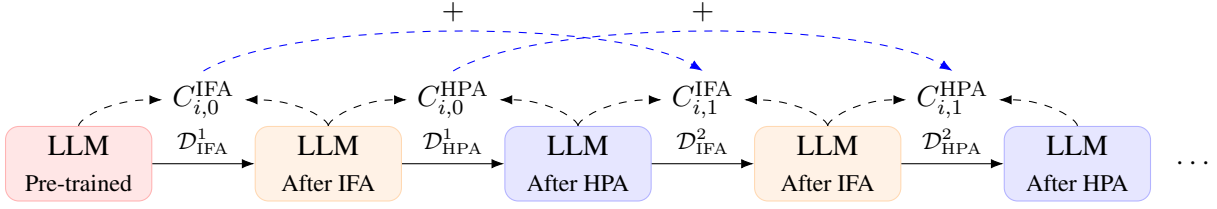
\begin{figure*}
    \centering
\begin{tikzpicture}
    \tikzstyle{lm} = [minimum width=5em, minimum height=5ex,rounded corners=5pt,align=center,];

    \node [draw=red!30,fill=red!10,lm, anchor=west] at (0, 0) (plm) {LLM \\ \small Pre-trained};
    \node [draw=orange!30,fill=orange!10, lm, anchor=west] at ([xshift=3.5em]plm.east) (ifa0) {LLM \\ \small After IFA};
    \node [draw=blue!30,fill=blue!10, lm, anchor=west] at ([xshift=3.5em]ifa0.east) (hpa0) {LLM \\ \small After HPA};
    \node [draw=orange!30,fill=orange!10, lm, anchor=west] at ([xshift=3.5em]hpa0.east) (ifa1) {LLM \\ \small After IFA};
    \node [draw=blue!30,fill=blue!10, lm, anchor=west] at ([xshift=3.5em]ifa1.east) (hpa1) {LLM \\ \small After HPA};
    \node [anchor=west] at ([xshift=.5em]hpa1.east) (dots) {$\cdots$};
    \draw [-Latex] (plm.east) -- (ifa0.west) node [midway, above] {\small $\mathcal{D}_\mathrm{IFA}^1$};
    \draw [-Latex] (ifa0.east) -- (hpa0.west) node [midway, above] {\small $\mathcal{D}_\mathrm{HPA}^1$};
    \draw [-Latex] (hpa0.east) -- (ifa1.west) node [midway, above] {\small $\mathcal{D}_\mathrm{IFA}^2$};
    \draw [-Latex] (ifa1.east) -- (hpa1.west) node [midway, above] {\small $\mathcal{D}_\mathrm{HPA}^2$};

    \node at ([yshift=2ex]$1/2*(plm.north)+1/2*(ifa0.north)$) (f0) {$C_{i,0}^\mathrm{IFA}$};
    \node at ($1/2*(ifa0.north)+1/2*(hpa0.north)$) (tmp0) {};
    \node at (tmp0|-f0) (f1) {$C_{i,0}^\mathrm{HPA}$};
    \node at ($1/2*(ifa1.north)+1/2*(hpa0.north)$) (tmp1) {};
    \node at (tmp1|-f0) (f2) {$C_{i,1}^\mathrm{IFA}$};
    \node at ($1/2*(ifa1.north)+1/2*(hpa1.north)$) (tmp2) {};
    \node at (tmp2|-f0) (f3) {$C_{i,1}^\mathrm{HPA}$};
    \draw [-Latex, dashed] (plm.north) .. controls ([yshift=.5ex]plm.north) and ([xshift=1em]plm.north|-f0.west) .. (f0.west);
    \draw [-Latex, dashed] (ifa0.north) .. controls ([yshift=.5ex]ifa0.north) and ([xshift=-1em]ifa0.north|-f0.east) .. (f0.east);
    \draw [-Latex, dashed] (ifa0.north) .. controls ([yshift=.5ex]ifa0.north) and ([xshift=1em]ifa0.north|-f1.west) .. (f1.west);
    \draw [-Latex, dashed] (hpa0.north) .. controls ([yshift=.5ex]hpa0.north) and ([xshift=-1em]hpa0.north|-f1.east) .. (f1.east);
    \draw [-Latex, dashed] (hpa0.north) .. controls ([yshift=.5ex]hpa0.north) and ([xshift=1em]hpa0.north|-f2.west) .. (f2.west);
    \draw [-Latex, dashed] (ifa1.north) .. controls ([yshift=.5ex]ifa1.north) and ([xshift=-1em]ifa1.north|-f2.east) .. (f2.east);
    \draw [-Latex, dashed] (ifa1.north) .. controls ([yshift=.5ex]ifa1.north) and ([xshift=1em]ifa1.north|-f3.west) .. (f3.west);
    \draw [-Latex, dashed] (hpa1.north) .. controls ([yshift=.5ex]hpa1.north) and ([xshift=-1em]hpa1.north|-f3.east) .. (f3.east);

    \draw [-Latex, dashed, draw=blue] (f0.north) .. controls ([yshift=4ex,xshift=3em]f0.north) and ([yshift=4ex,xshift=-3em]f2.north) .. (f2.north) node [midway, above] {$+$};
    \draw [-Latex, dashed, draw=blue] (f1.north) .. controls ([yshift=4ex,xshift=3em]f1.north) and ([yshift=4ex,xshift=-3em]f3.north) .. (f3.north) node [midway, above] {$+$};
\end{tikzpicture}
    \vspace{-5mm}
    \caption{Architecture of \textsc{Hbat}. 
    We introduce the alternating alignment and the modified EWC methods to design \textsc{Hbat}, which enables it to address optimization conflict problem in the process of LLM alignment training. Here, black solid arrows (\tikz[baseline=-0.5ex]\draw[-Latex] (0,0) -- (0.7,0);) denote learning from the subsets $\mathcal{D}_\mathrm{IFA}^{n}$ and $\mathcal{D}_\mathrm{HPA}^{n}$ via Eq. \ref{eq:ifa} and Eq. \ref{eq:hpa}, respectively. 
    Black dashed arrows (\tikz[baseline=-0.5ex]\draw[-Latex, dashed] (0,0) -- (0.7,0);) denote computing the amount of parameter changes before and after training and blue dashed arrows (\tikz[baseline=-0.5ex]\draw[-Latex, dashed, draw=blue] (0,0) -- (0.7,0);) denote accumulating the parameter changes resulting from learning all previous subsets (see Section \ref{sec:alternationAlignmnet}).
    \textbf{IFA}: instruction-following alignment;
    \textbf{HPA}: human-preference alignment.
    }
    \label{fig:mian_figure}
    \vspace{-4mm}
\end{figure*}

\section{Background}
Despite the extensive knowledge endowed from pre-training, LLMs are difficult to produce content that humans want.
This is because that pre-trained LLMs lack understanding of input instructions and human preferences.
To address this, we often perform alignment training on them, first for instruction-following alignment and then for human-preference alignment.

\subsection{Instruction-Following Alignment}
Instruction-following alignment enables the pre-trained language model to  acquire the capability to understand and follow instructions in the prompt by mimicking the human-labeled response.
Specifically, given a human prompt $x$ and the labeled response of $N_{y}$ tokens $y = \{y_{1},\dots, y_{N_{y}}\}$, where each token $y_{t}$ is drawn from a vocabulary.
In the training process, the LLM learns the probability:
\begin{eqnarray}
    p_{\theta}(y|x) = \prod_{t=1}^{N_{y}}p_{\theta}(y_{t}|y_{<t},x)
    \label{eq:generativeProb}
\end{eqnarray}
where $y_{<t}$ is the prefix $\left\lbrace y_{1}, y_{2}, \dots, y_{t-1}\right\rbrace $, and $\theta$ is a trained parameter set.
The standard training objective is to maximize the likelihood over all the tokens of the labeled response, \textit{i.e., maximum likelihood estimation (MLE)} \cite{myung2003tutorial}.
The corresponding loss function can be defined by:
\begin{eqnarray}
    \mathcal{L}_{\mathrm{MLE}}=-\sum_{t}\log p_{\theta}(y_{t}|y_{<t},x)
    \label{eq:mle}
\end{eqnarray}

\subsection{Human-Preference Alignment}
This process of human-preference alignment consists of two main steps: 1) learning a preference model from comparison response pairs to act as a reward model, and 2) maximizing the reward, written as $\arg\max_{\theta}\mathbb{E}p_{\theta}(\hat{y}|x)[r(\hat{y})]$, where $\hat{y}$ is a generated response and $r(\cdot)$ denotes the computation of the reward for $\hat{y}$ using a reward model.
We usually employ an RL algorithm to achieve step 2.
Taking PPO as an instance, the corresponding loss for this training sample is given by:
\begin{eqnarray}
\begin{aligned}
    \mathcal{L}_{\mathrm{PPO}}=&-\sum_{\hat{y} \in \Omega(x)}\log p_{\theta}(\hat{y}|x) r(\hat{y})
    \\&-\alpha\log(\frac{p_{\theta}(\hat{y}|x)}{p_{\theta_{old}}(\hat{y}|x)})
\end{aligned}
\end{eqnarray}
where $\Omega(x)$ is the output space which comprises all possible responses for prompt $x$, $\theta_{old}$ is the parameter set of the LLM trained via instruction-following alignment, and $\alpha$ is a KL reward coefficient which controls the strength of the KL penalty $\log(\frac{p_{\theta}(\hat{y}|x)}{p_{\theta_{old}}(\hat{y}|x)})$. 
Here, $\Omega(x)$ is approximated using the Monte Carlo method \cite{williams1992simple}.

To bypass the complex RL procedure, \citet{rafailov2023direct} proposed DPO method, which employs a reward model training objective to maximize rewards.
It gives a new loss function:
\begin{eqnarray}
    \begin{aligned}
        \mathcal{L}_{\mathrm{DPO}}=&-\log \sigma [ \beta \log(\frac{p_{\theta}(y_{w}|x)}{p_{\theta_{old}}(y_{w}|x)})\\&-\beta \log(\frac{p_{\theta}(y_{l}|x)}{p_{\theta_{old}}(y_{l}|x)})]
    \end{aligned}
\end{eqnarray}
where $(y_{w}, y_{l})$ is two of the different responses and $y_{w}$ aligns better with human preferences than $y_{l}$.
$\beta$ is a scaling factor and $\sigma$ is a Sigmoid function.

\section{Method}
In this work, we aim to solve an optimization conflict limitation during alignment training.
We propose the \textsc{Hbat} to achieve this.
The overview of \textsc{Hbat} is depicted in Figure \ref{fig:mian_figure}.
As shown in the figure, we propose the alternating alignment and modified EWC in \textsc{Hbat} to achieve our goal.
In the following subsections, we will describe them.

\subsection{Alternating Alignment}
\label{sec:alternationAlignmnet}
We first introduce the optimization conflict problem in the alignment training.
Suppose that we have training datasets $\mathcal{D}_\mathrm{IFA}$ and $\mathcal{D}_\mathrm{HPA}$ for instruction-following alignment and human-preference alignment, respectively.
We expect that the LLM will simultaneously align instructions and human preferences well by learning from both datasets.
However, during the traditional two-stage alignment training, while the LLM learns from new training samples in $\mathcal{D}_\mathrm{HPA}$, it may have conflicts with previous knowledge learned from $\mathcal{D}_\mathrm{IFA}$.

Inspired by the success of interactive methods in multi-objective optimization, we propose an alternating alignment method.
In the alternating alignment, we redesign the relationship between the instruction-following alignment and human-preference alignment to offer a refinement of the collaboration among them.
Specifically, we divide the datasets $\mathcal{D}_\mathrm{IFA}$ and $\mathcal{D}_\mathrm{HPA}$ into $N$ mutually exclusive splits $\{\mathcal{D}_\mathrm{IFA}^1, \mathcal{D}_\mathrm{IFA}^2, \cdots, \mathcal{D}_\mathrm{IFA}^{N}\}$ and $\{\mathcal{D}_\mathrm{HPA}^1, \mathcal{D}_\mathrm{HPA}^2, \cdots, \mathcal{D}_\mathrm{HPA}^N\}$, respectively.
The LLM performs an alternating alignment by sequentially learning from $\{\mathcal{D}_\mathrm{IFA}^1, \mathcal{D}_\mathrm{HPA}^1, \cdots, \mathcal{D}_\mathrm{HPA}^N\}$.
In each round of alternate training, the human-preference alignment acts as a “decision maker” to offer preference information.
This preference information enables an LLM to align human preferences following instruction alignment.

\subsection{Elastic Weight Consolidation}
\label{sec:ewc}
To further solve the optimization conflict, we introduce a modified EWC to alternating alignment.
Firstly, we add EWC to the process of human-preference alignment to mitigate optimization conflicts with instruction-following alignment.
The loss of human-preference alignment with EWC is:
\begin{eqnarray}
    \mathcal{L}_{\mathrm{HPA}}=\mathcal{L}_{\mathrm{PPO}}+\sum_{i}\frac{\lambda}{2}F_{i}^\mathrm{IFA}(\theta_{i}-\theta^\mathrm{IFA}_{i})^2
    \label{eq:hpa}
\end{eqnarray}
where $i$ is the index corresponding to each parameter within the LLM, $\theta^\mathrm{IFA}$ is the parameter set of the LLM trained by instruction-following alignment, $\lambda$ is a balance factor, and 
$F$ is the diagonal of the empirical Fisher matrix \cite{pascanu2013revisiting}.
Here, $F_{i}^\mathrm{IFA}$ denotes how important the $i$-th parameter $\theta^\mathrm{IFA}_{i}$ is to the instruction-following alignment.
Note that we can replace $\mathcal{L}_{\mathrm{PPO}}$ with other loss functions, such as $\mathcal{L}_{\mathrm{DPO}}$, which can align LLMs with human preferences.

\paragraph{Modified EWC for LLMs.} 
However, the original EWC introduces a large computational overhead on the alignment training.
This is because estimating $F_{i}^\mathrm{IFA}$ requires the LLM to be additionally trained multiple times on the whole training set (see Appendix \ref{appendix:fim}).
To mitigate this problem, we redesign this estimation approach, and use the amount of parameter changes before and after model training to compute the $F$.
Furthermore, considering that LLMs typically have a large number of parameters and the size of the $F$ will be enormous, we attempt to implement EWC at the granularity of parameter units.
Specifically, we redefine $F$ as a numerical value, with $F_{i}^\mathrm{IFA}$ representing how importance of the parameter unit $\theta_{i}^\mathrm{IFA}$ as a whole to the instruction-following alignment.
This redefined $F$ can be given by:
\begin{eqnarray}
    F_{i}^\mathrm{IFA} = F_{max}\times \frac{e^{C_{i}^\mathrm{IFA}}}{\sum_{i}e^{C_{i}^\mathrm{IFA}}}
    \label{eq:computeF}
\end{eqnarray}
where $F_{max}$ is the maximum value of $F$.
$C_{i}^\mathrm{IFA}$ denotes the amount of parameter $\theta_{i}$ changes before and after instruction-following alignment training for the LLM, written as:
\begin{eqnarray}
    C_{i}^\mathrm{IFA}=\frac{1}{|\theta_{i}|}\sum_{j=1}^{|\theta_{i}|}(\theta_{i,j}^{before}-\theta_{i,j}^\mathrm{IFA})^2
    \label{eq:computeC}
\end{eqnarray}
where $j$ is the index corresponding to each neuron within a parameter, $|\theta_{i}|$ is the number of neurons contained in the parameter $\theta_{i}$, and $\theta^{before}$ is the parameter set of the LLM before instruction-following alignment training.

\begin{algorithm}[t]
\caption{Hybrid Alignment Training}
\begin{algorithmic}[1]
    \INPUT the pre-trained LLM $\mathcal{M}$; the instruction-following alignment training dataset $\mathcal{D}_\mathrm{IFA}$; the human-preference alignment training dataset $\mathcal{D}_\mathrm{HPA}$
    \OUTPUT the aligned LLM $\mathcal{M}$;
    \State divide $\mathcal{D}_\mathrm{IFA}$ and $\mathcal{D}_\mathrm{HPA}$ into $N$ subsets respectively;\label{alg:line1}
        \For{$n=1$ to $N$}
        \label{alg:line3}
            \If{$n$==1}
                \State train $\mathcal{M}$ on first subset of $\mathcal{D}_\mathrm{IFA}$ via Eq. \ref{eq:mle};
            \Else
                \State compute the $F^\mathrm{HPA}$ via Eq. \ref{eq:fhpa};
                \State train $\mathcal{M}$ on $n$-th subset of $\mathcal{D}_\mathrm{IFA}$ via Eq. \ref{eq:ifa};
            \EndIf
            \State compute the $F^\mathrm{IFA}$ via Eq. \ref{eq:computeF};
            \State train $\mathcal{M}$ on $n$-th subset of $\mathcal{D}_\mathrm{HPA}$ via Eq. \ref{eq:hpa};
        \EndFor \label{alg:line11}
    \State return $\mathcal{M}$
\end{algorithmic}

\label{alg:hat}
\end{algorithm}

\subsection{EWC for Alternating Alignment}
We apply EWC on a global scale during alternating alignment.
Specifically, we add the modified EWC not only when learning each divided subset from $\mathcal{D}_\mathrm{HPA}$ as described in Section \ref{sec:ewc}, but also when learning each divided subset from $\mathcal{D}_\mathrm{IFA}$.
The motivation is that the instruction-following alignment can likewise lead to an optimization conflict with human-preference alignment.
$\mathcal{L}_{\mathrm{IFA}}$ can be induced by:
\begin{eqnarray}
\mathcal{L}_{\mathrm{IFA}}=\mathcal{L}_{\mathrm{MLE}}+\sum_{i}\frac{\lambda}{2}F_{i}^\mathrm{HPA}(\theta_{i}-\theta^\mathrm{HPA}_{i})^2
\label{eq:ifa}
\end{eqnarray}
where $\theta^\mathrm{HPA}$ is the parameters of the LLM trained by human-preference alignment.
Here, similar to $F_{i}^\mathrm{IFA}$, $F_{i}^\mathrm{HPA}$ can be computed by: 
\begin{eqnarray}
F_{i}^\mathrm{HPA}=F_{max} \times \frac{e^{C_{i}^\mathrm{HPA}}}{\sum_{i}e^{C_{i}^\mathrm{HPA}}}
\label{eq:fhpa}
\end{eqnarray}
where $C_{i}^\mathrm{HPA}$ denotes the amount of parameter $\theta_{i}$ changes before and after human-preference alignment training for the LLM.
It can be computed via Eq. \ref{eq:computeC}.
Note that when learning the first subset $\mathcal{D}^1_\mathrm{IFA}$, since the LLM has not yet been trained with human preferences, we only employ the $\mathcal{L}_{\mathrm{MLE}}$.

In the process of alternating alignment training, learning a new subset from one alignment training dataset can produce optimization conflicts.
These conflicts arise not only with the closest subset from another alignment training dataset but also with all the previous subsets within this dataset.
Thus, when estimating $F$, we consider the parameter changes resulting from all previous subsets in another alignment training dataset.
To this end, we replace the $C_{i}^\mathrm{IFA}$ and $C_{i}^\mathrm{HPA}$ in Eqs. \ref{eq:ifa} and \ref{eq:hpa} with accumulated parameter changes $AC_{i}^\mathrm{IFA}$ and $AC_{i}^\mathrm{HPA}$ from all previous subsets in $\mathcal{D}_\mathrm{IFA}$ and $\mathcal{D}_\mathrm{HPA}$, respectively.
Here, when learning from $n$-th subset, we compute $AC_{i,n}^\mathrm{IFA}$ and $AC_{i,n}^\mathrm{HPA}$ by:
\begin{eqnarray}
    AC_{i,n}^\mathrm{IFA}=\sum_{k=1}^{n}C_{i,k}^\mathrm{IFA}
    ,
    AC_{i,n}^\mathrm{HPA}=\sum_{k=1}^{n}C_{i,k}^\mathrm{HPA}
\end{eqnarray}
where $C_{i,k}^\mathrm{IFA}$ and $C_{i,k}^\mathrm{HPA}$ are the amount of parameter changes produced at learning $k$-th subset in $\mathcal{D}_\mathrm{IFA}$ and $\mathcal{D}_\mathrm{HPA}$, respectively.
The process of our \textsc{Hbat} is also described in Algorithm \ref{alg:hat}.

\section{Experimental Setup}
We evaluated \textsc{Hbat} on summarization and dialogue tasks based on the commonly used LLaMA2-7B and LLaMA2-13B models. 
\subsection{Datasets}
The datasets used for each task are as follows:
\paragraph{Summarization.}
We used the same dataset as \citet{stiennon2020learning}, which is a filtered version\footnote{\url{https://github.com/openai/summarize-from-feedback}} of the TL;DR dataset \cite{volske2017tl}.
The filtered training set consists of 120k Reddit posts with accompanying summaries.
For instruction-following training and human-preference alignment training, we used all posts in a filtered training set, respectively.
The filtered test set and validation set contain 6,553 posts and 6,447 posts respectively, which would result in a huge computational cost when used on a large scale.
Thus, we randomly selected 10\% of posts from them as a test set and a validation set in our experiments, respectively.
For training reward models, we employed the open-source 92.9k summary comparisons\footnote{\url{https://huggingface.co/datasets/openai/summarize_from_feedback}}.

\paragraph{Dialogue.}
We conducted experiments on the Alpaca data \cite{taori2023alpaca} which contains 52k training samples.
Here, we employed the sliced data splits\footnote{\url{https://huggingface.co/datasets/tatsu-lab/alpaca_farm}} released by AlpacaFarm \cite{dubois2023alpacafarm} to conduct instruction-following alignment training, reward model training, and human-preference alignment training.
Note that we used the human preferences rather than the simulated preferences to train our reward models.
In the evaluation, we employed the AlpacaFarm evaluation set which consists of 805 instructions.
We randomly selected 200 instructions from them as our validation set and the rest as our test set.


\subsection{Settings}
We trained reward models with the ranking loss for all tasks, following \citet{stiennon2020learning}.
For instruction-following alignment training, we employed the cross-entropy loss on batches of prompts concatenated with responses, computing the loss only on the response tokens.
For human-preference alignment training, we used PPO and DPO as our base algorithms.
For \textsc{Hbat}, we set the number of dataset splits to 2 and 10 for dialogue and summarization tasks, respectively.
Additionally, we employed a top-$p$ sampling strategy for generation, where the temperature and $p$ were set to 0.75 and 0.95, respectively, values that are commonly used in real-world applications.
We publicly release all our code used for the experiments described in this work\footnote{\url{https://github.com/wangclnlp/DeepSpeed-Chat-Extension/tree/main/examples/hybrid_alignment_training}}.
More training details are shown in Appendix \ref{sec:experimental_detail}.

\begin{table*}[t]
    \centering
    \scalebox{0.88}{
    \begin{tabular}{lrccccccccc}
\toprule[1.1pt]
\multirow{2}{*}{\textbf{Method}} & \multirow{2}{*}{\textbf{\#Param}} & \multirow{2}{*}{\textbf{PPO}} & \multirow{2}{*}{\textbf{DPO}}& \multicolumn{4}{c}{\textbf{Summarization}}   & \multicolumn{3}{c}{\textbf{Dialogue}}  \\ \cmidrule(l){5-8} \cmidrule(l){9-11} 
& & & &\multicolumn{1}{c}{ROUGE-L} & \multicolumn{1}{c}{BS} & \multicolumn{1}{c}{Reward} & \multicolumn{1}{c}{Win} & \multicolumn{1}{c}{PandaLM} & \multicolumn{1}{c}{Reward} & \multicolumn{1}{c}{Win} \\ \midrule
\multicolumn{11}{l}{\textit{\textbf{Based on LLaMA2-7B Model}}} \\
\midrule
SFT        &7B & &  & 22.60   & -5.46   & 3.72   & 53.20 & 54.76  & -6.79 & 43.49    \\ \midrule
RLHF    &7B &\checkmark &  & 25.85   & -4.27   & 4.43   & 63.80 &69.79   &-5.81 & 55.63     \\
RLHF+pt &7B &\checkmark &  & 22.25   & -5.64   & 3.74   & 56.26 & 53.52  & -7.09 & 54.18     \\ 
SFT+ppo &7B &\checkmark &  & 13.75   & -5.78   & 2.40   & 18.91 &45.32 &-8.60 & 42.25     \\ 
\textsc{Hbat}-Freeze &7B &\checkmark & & 25.33   & -4.28   & 5.26   & 64.79 & 69.91 & -5.91 & 56.19    \\
\textsc{Hbat} (Ours) &7B &\checkmark &  & \bf26.18   & \bf-3.82   & \bf5.74   & \bf72.52 & \bf70.88  & \bf-5.37 & \bf57.12 \\ \midrule
DPO  &7B & &\checkmark  & 22.96   & -5.13   & 4.27   & 61.37 & 70.74  & -5.72 &   54.23 \\
\textsc{Hbat}-Freeze &7B & &\checkmark  &23.01 &-5.05 &4.45 &64.18 &68.78  & -5.41 &56.95    \\
\textsc{Hbat} (Ours) &7B & &\checkmark  &\bf23.14    &\bf-4.18    &\bf4.95    &\bf70.58   & \bf74.78  & \bf-5.22  & \bf58.10  \\
\midrule
\multicolumn{9}{l}{\textit{\textbf{Based on LLaMA2-13B Model}}} \\
\midrule
SFT  &13B & & & 23.27   & -5.12   & 4.01   & 57.91 & 62.16  & -6.32 & 46.11  \\ \midrule
RLHF &13B &\checkmark  &   & 24.51   & -3.96   & 5.55   & 71.67 & 72.21  & -5.65 &  61.16 \\
RLHF+pt &13B &\checkmark &  & 22.92   & -5.49   & 3.97   & 64.42 &   63.67 &-6.97 & 54.45     \\ 
SFT+ppo &13B &\checkmark & & 13.84   & -5.97   & 2.53   & 28.97 & 54.00   &-7.93 & 43.12   \\  
\textsc{Hbat}-Freeze &13B &\checkmark & & 25.80   & -3.63   & 6.18   & 77.22 & 71.31 & -5.49 & 56.37    \\
\textsc{Hbat} (Ours) &13B &\checkmark & & \bf26.77   & \bf-3.51   & \bf6.41   & \bf78.81 & \bf72.83 & \bf-5.11  & \bf62.32   \\ \midrule
DPO    &13B & &\checkmark   & 23.02   & -5.39   & 4.55   & 69.40 & 75.00  & -5.07 & 64.31  \\
\textsc{Hbat}-Freeze &13B & &\checkmark  &23.10 &-5.08 & 4.85 &71.44 &76.87  & -5.01 & 65.62    \\
\textsc{Hbat} (Ours) &13B & &\checkmark &\bf24.12    &\bf-4.05    &\bf5.40    &\bf74.92  & \bf77.79 & \bf-4.78  &\bf67.45   \\ 

\bottomrule[1.1pt]
\end{tabular}}
    \caption{
    Results on summarization and dialogue tasks.
    The best results for each group are in \textbf{bold}.
    The “BS” and “Win” columns report the BARTScore and the win rate as assessed by GPT-4, respectively.
    The “PPO” and “DPO” columns denote that we employ PPO and DPO during human-preference alignment training, respectively.
    }
    \label{tab:Res}
    \vspace{-4mm}
\end{table*}

\subsection{Evaluation Metrics}
For the summarization task, we measured the summary quality by computing ROUGE \cite{lin2004rouge} and BARTScore \cite{yuan2021bartscore}, respectively.
For the dialogue task, we measured the response quality with PandaLM \cite{wang2023pandalm} which can distinguish the superior model from some LLMs.
To further evaluate the performance of the model, we employed GPT-4 as a proxy for human evaluation of summary and response quality in the dialogue and summarization tasks, where the used evaluation prompts were the same as in \citet{rafailov2023direct}.
We used reference summaries and responses in the test set as the baseline.
Additionally, following \citet{stiennon2020learning}'s work, we evaluated the model by computing the reward scores of test sets via our reward models.

\subsection{Baselines}
Our baselines are the standard two-stage alignment training (referred to as \textbf{RLHF/DPO}) and the commonly used instruction-following alignment training (referred to as \textbf{SFT}).
Furthermore, we compared the proposed \textsc{Hbat} with commonly used multi-objective optimization methods, including adding a pre-training loss in the human-preference alignment training (\textbf{RLHF+pt}) \cite{ouyang2022training} and adding a human-preference alignment loss in the instruction-following alignment training (\textbf{SFT+ppo}) \cite{wang2023learning}.
To evaluate the effectiveness of EWC, we also chose the \textbf{\textsc{Hbat}-Freeze} method as a baseline, where we directly froze important parameters instead of EWC.

\begin{table*}[t]
\centering
\scalebox{0.77}{

\begin{tabular}{lccccccccccc}
\toprule[1.1pt]
&  &  & \multicolumn{4}{c}{Summarization}    & \multicolumn{5}{c}{Dialogue}  \\ \cmidrule(l){4-7}  \cmidrule(l){8-12} 
\multirow{-2}{*}{Method} & \multirow{-2}{*}{PPO} & \multirow{-2}{*}{DPO} & Coherence  & Accuracy      & Coverage    & Overall  & Fluency  & Accuracy  & Toxicity & Helpfulness  & Overall \\ \midrule
SFT     &   & & 5.63 & 4.91 & 5.03  & 5.13 & 8.84 & 7.77         & 8.49  & 7.43 & 7.31  \\ \midrule
RLHF   &\checkmark & & 5.84 & 4.63 & 4.82  & 5.16         & 8.39         & 7.62         & \textbf{8.87}& \textbf{7.47}& 7.72         \\ 
\textsc{Hbat}  &\checkmark    &    & \textbf{6.23}& \textbf{4.83}& \textbf{5.77}& \textbf{5.69}& \textbf{8.80}& \textbf{7.70}& 8.48         & 7.39         & \textbf{7.89} \\ \midrule
DPO   &  &\checkmark   & 5.83         & 4.45         & 5.20         & 5.27         & 8.67         & 7.13         & \textbf{8.54}& \textbf{7.63}& 7.84         \\ 
\textsc{Hbat} &    &\checkmark     & \textbf{5.93}& \textbf{5.01}& \textbf{5.40}& \textbf{5.49}& \textbf{8.79}& \textbf{7.80}& 8.45         & 7.51         & \textbf{7.96}\\ \bottomrule[1.1pt]
\end{tabular}}
\caption{The results of human evaluation on the LLaMA2-13B model for our \textsc{Hbat} and baselines.}
\vspace{-5mm}
\label{tab:human-evaluation}
\end{table*}

\subsection{Experimental Results}
Table \ref{tab:Res} displays the experimental results on summarization and dialogue tasks.
\vspace{-1mm}
\paragraph{Results of Summarization.}
First, compared with the traditional two-stage alignment training and instruction-following alignment training, the proposed \textsc{Hbat} can achieve optimal results on both of LLaMA2-7B and LLaMA2-13B.
Notably, \textsc{Hbat} outperforms RLHF by 7.14 points on the GPT-4 win rate when using PPO on the LLaMA2-13B model.
Second, compared with multi-task learning-based methods, including RLHF+pt and SFT+ppo, we can see that \textsc{Hbat} has significant improvements on all evaluation metrics.
For instance, compared to RLHF+pt, \textsc{Hbat} yields a +3.93 ROUGE-L improvement on the LLaMA2-7B model.
Also, we see that the multi-objective optimization method can hurt alignment, \textit{e.g.,} RLHF+pt loses 0.69 Reward points on the LLaMA2-7B model.
The phenomenon aligns with the observation reported in \citet{ouyang2022training}'s work.
One potential explanation can be that while these multi-objective optimization methods achieve optimization of these objectives simultaneously, they still suffer from serious optimization conflict \cite{zhang2021survey}.
Third, when using DPO during human-preference alignment training, our \textsc{Hbat} is consistently better than all baselines.
For a LLaMA2-13B model, it obtains a GPT-4 win rate of 74.92.

\vspace{-3mm}
\paragraph{Results of Dialogue.}
We also evaluate the proposed \textsc{Hbat} on the dialogue task. 
Similarly, when using PPO during human-preference alignment training, we can observe that \textsc{Hbat} outperforms RLHF by a large margin (\textit{e.g.,} 2.21 PandaLM and 0.54 Reward benefits on the LLaMA2-13B model).
However, different from the summarization task, we find that DPO can achieve better performance than PPO on the dialogue task.
For instance, when using LLaMA2-13B, \textsc{Hbat} with DPO can outperform PPO by a margin of 5.13 points on the GPT-4 win rate.
We assume that this is attributed to the reward model quality. 
To verify this assumption, we conduct tests on the employed reward models and find a significant difference in accuracy between the two tasks: the accuracy of the reward model for the summarization task significantly exceeds that of the dialogue task, achieving 0.75 compared to 0.65, respectively.

Furthermore, compared with \textsc{Hbat}-Freeze, we see that \textsc{Hbat} achieves better performance on all tasks.
It demonstrates that freezing specific parameters is inferior to constraining specific parameters.
We attribute this to the fact that the freezing operation reduces the amount of learnable parameters, which imposes a hurdle to learning new knowledge.

\subsection{Human Evaluation}
We further conduct a human evaluation of the obtained results through comprehensive evaluation aspects.
For the summarization task, following \citet{stiennon2020learning}, we consider four evaluation aspects, including coherence, accuracy, coverage, and overall score.
We provide three optional scores of 1, 4, and 7 for each evaluation aspect. 
Similarly, for the dialogue task, we consider five evaluation aspects: fluency, accuracy, toxicity, helpfulness, and overall score. 
We have defined detailed evaluation rubrics similar to those for the summarization task. Please refer to Table \ref{tab:rubric_human_evaluation} in the Appendix for descriptions of all the evaluation rubrics.
The results of human evaluation on the LLaMA2-13B model are shown in Table \ref{tab:human-evaluation}. 
From these evaluation results, we see that whether using PPO or DPO, our \textsc{Hbat} can achieve optimal results compared to the baselines in terms of the overall score of human evaluation. 
This demonstrates that \textsc{Hbat} has a great advantage in practical NLP applications. 
Furthermore, we also see that in some evaluation aspects, our \textsc{Hbat} is weaker compared to the baseline, \textit{e.g.}, helpfulness aspect on the dialogue task. 
This is due to the fact that the aim of our \textsc{Hbat} is to achieve a Pareto alignment between the instruction-following alignment and the human-preference alignment, not to improve for a particular alignment.

\begin{table*}[h]
    \centering
    \scalebox{0.72}{
    \begin{tabular}{ll}
\toprule[1.1pt]
\multicolumn{2}{l}{\textit{\textbf{Case 1}}} \\ \midrule
Input & \parbox{20cm}{Please write the Excel function name associated with each description.- Returns the number of days between two dates - Returns the starting position of a text string within another text string. - Returns the number in the middle of the set of given numbers.} \\ \cmidrule(l){2-2} 
SFT   & \parbox{19cm}{1. DAYS(date1, date2) 2. FIND(text1, text2) 3. MID(text1, start\_num, num).\\
\bf{Fluency:7  Accuracy:6 Toxicity:10 Helpful:5 Overall:6}} \\[4ex]
RLHF  & \parbox{20cm}{1.=DATEDIF(A2,B2,"d") 2.=FIND(A2,B2) 3.=MID(A2:B2,FIND(A2:B2,A2)+1,1) 4.=RANDBETWEEN(A2,B2) 5.=SUMIF(A2:B2,">="\&A2,A2:B2) 6.=SUMPRODUCT(A2:B2) 7.=SUM(A2:B2) 8.=AVERAGE(A2:B2) 9.=AVERAGEIF(A2:B2,"<"\&A2,A2:B2) 10.=COUNTIF(A2:B2,">"\&A2) 11.=COUNTIFS(A2:B2,"<"\&A2,A2:B2,">="\&B2,A2:B2,"<="\&B2) 12.=IFERROR(VLOOKUP(A2,A2:B2,2,0),"")...\\
\bf{Fluency:5  Accuracy:4 Toxicity:10 Helpful:6 Overall:4}
} \\[8ex]
\textsc{Hbat}  & \parbox{20cm}{1.DATEDIF(\textbackslash em\{date1\},\} date2\textbackslash em\{,\} \textbackslash em\{“day” \textbackslash em\{) \textbackslash n2. FIND(\textbackslash em\{text1\},\} text2\textbackslash em\{,\} \textbackslash em\{\} \textbackslash em\{\} \textbackslash em\{\} \textbackslash em\{\} \textbackslash n3. MID(\textbackslash em\{text\},\} start\textbackslash em\{,\} numchars\textbackslash em\{)\} \\
\bf{Fluency:7  Accuracy:6 Toxicity:10 Helpful:6 Overall:8}
}  \\ \midrule
\multicolumn{2}{l}{\textit{\textbf{Case 2}}} \\ \midrule
Input & \parbox{20cm}{I assume you are familiar with the Drake equation. Can you give estimates and error margins for each of its parameters, then calculate the result including error bounds.} \\  \cmidrule(l){2-2} 
SFT   & \parbox{20cm}{The number of stars in the galaxy is estimated to be between $10^{22}$ and $10^{24}$, with a mid-range value of $10^{23}$. \textbackslash n2. The fraction of stars that have planets is estimated to be between 0.1 and 0.2, with a mid-range value of 0.15. \textbackslash n3. The average number of planets that can support life per star is estimated to be between 1 and 2, with a mid-range value of 1.5. \textbackslash n4. The fraction of planets where life has evolved is estimated to be between 0.01 and 0.1, with a mid-range value of 0.05. \textbackslash n5. The fraction of planets where life has evolved that develops intelligent life is estimated to be between 0.01 and 0.05, with a mid-range value of 0.03...\\
\bf{Fluency:9  Accuracy:6 Toxicity:8 Helpful:6 Overall:5}
} \\[10ex]
RLHF  & \parbox{20cm}{
1. Number of civilizations in our Galaxy: The number of civilizations in our Galaxy can be estimated using the Drake equation. The equation states that the number of civilizations in the Galaxy (N) is equal to the product of the rate of star formation (R*) multiplied by the fraction of stars with planets (fP) multiplied by the fraction of planets that can support life (fL) multiplied by the fraction of planets with life (fL) multiplied by the fraction of planets with intelligent life (fI) multiplied by the fraction of intelligent civilizations that are capable of communicating with us (fC) multiplied by the fraction of those that are willing to communicate (fW): \textbackslash n\textbackslash nN = R* fP fL fI fC fW\textbackslash n \textbackslash nThe rate of star formation (R*) can be estimated to be 1-2 stars per year in our Galaxy... \\
\bf{Fluency:5  Accuracy:5 Toxicity:8 Helpful:8 Overall:7}
} \\[11ex]
\textsc{Hbat}  & \parbox{20cm}{
1. The number of stars with planetary systems (N*): Unknown, but estimated to be between $10^{-5}$ and $10^{-8}$ per star. Error margin: 10 orders of magnitude. \textbackslash n2. The fraction of stars with planetary systems that could support life (fp): Unknown, but estimated to be between $10^{-5}$ and 0.5. Error margin: 10 orders of magnitude. \textbackslash n3. The average number of planets that can support life per star with a planetary system (ne): Unknown, but estimated to be between 1 and 10. Error margin: 1 order of magnitude. \textbackslash n4. The fraction of planets that could support life that actually develop life (fl): Unknown, but estimated to be between 0.01 and 0.99. Error margin: 1 order of magnitude. \textbackslash n5. The ... \\
\bf{Fluency:7  Accuracy:6 Toxicity:8 Helpful:8 Overall:8}
} \\
\bottomrule[1.1pt]
\end{tabular}}
    \caption{Several cases from the dialogue task on the LLaMA2-13B model.}
    \label{tab:case_study}
    \vspace{-2mm}
\end{table*}

Table \ref{tab:case_study} presents several cases for human evaluation.
Case 1 shows that RLHF (\textit{i.e.}, human-preference alignment) won't always improve the performance of LLM trained by SFT (\textit{i.e.}, instruction-following alignment). 
This demonstrates that these two alignment optimization objectives are different, and aligning LLMs with these objectives in sequence might cause an optimization conflict. 
For example, in this case, SFT is probably more concerned with instruction following and response accuracy, while RLHF is more concerned with response helpfulness. 
In Case 2, we can observe that although the overall score of RLHF has increased, the fluency and accuracy scores have decreased, compared to SFT. 
In this case, our \textsc{Hbat} achieves a Pareto alignment through iterative alignment and modified elastic weight consolidation approaches, which aim to find a relatively optimal trade-off between instruction-following and human-preference alignment, thus achieving a better performance. 

\begin{table}[t]
    \centering
    \scalebox{0.80}{
    \centering
\begin{tabular}{lccc}
\toprule[1.1pt]
Method & PandaLM & Reward & Win \\ \midrule
SFT            &43.64 & -6.80 & 43.08    \\ 
DPO            &69.97 & -5.68 & 53.80    \\  \midrule
\textsc{Hbat}            &\bf75.76 &\bf-5.11 &\bf60.10    \\  
w/o EWC \ \ \ \ \ \ \ \ \  \ \ &67.53  &-5.76  &54.75  \\ 
w/o Alternating Alignment & 70.50 & -5.26 & 56.92    \\
\bottomrule[1.1pt]
\end{tabular}}
    \caption{
        Ablation studies on the components of \textsc{Hbat}.
        We report the scores for the dialogue validation set.
    }
    \label{tab:ablationStudy}
    \vspace{-5mm}
\end{table}


\subsection{Ablation Studies}
In this section, we present detailed ablation studies to explore the effects of EWC and alternating alignment with DPO on the LLaMA2-7B model.
The experiments are conducted on the dialogue dataset, and the impacts of removing each method are thoroughly examined.
The results are summarized in Table \ref{tab:ablationStudy}.
From the results, we see that the modified EWC can significantly improve response quality.
Notably, \textsc{Hbat} obtains a +5.35 points improvement on GPT-4 win rate with the modified EWC.
Additionally, the results indicate a significant dependency of our \textsc{Hbat} on the alternating alignment. 
The absence of this method results in \textsc{Hbat} fails a well-performed dialogue model.

\subsection{Analysis}

\begin{figure}
    \centering
    \tikzstyle{every node}=[scale=0.95]
\begin{tikzpicture}
    \matrix (m) [
    fill=white,
    draw=black,
    at={(0,3.6ex)},
    anchor=north west,
    cells={anchor=west},
    inner sep=1.8pt]
    {
        &[2.8em] \LegendImage{color=myred,mark=pentagon*,mark size=2pt} &  \LegendEntry{\small DPO}; \
        &[3.4em] \LegendImage{color=myblue,mark=diamond*,mark size=2pt} &  \LegendEntry{\small \textsc{Hbat}}; &[2.8em]\\
    };	
    \scriptsize{
        \begin{axis}
            [ymajorgrids,
            xmajorgrids,
            grid style=dashed,
            anchor=north west,
            at={(0,0)},
            width=.26\textwidth,
            symbolic x coords={1,2,3,4,5},
            xmax=5,
            xmin=1,
            ymin=-6.5,
            ymax=-4.8,
            xtick=data,
            x tick label style={/pgf/number format/fixed,
                /pgf/number format/fixed zerofill,
            /pgf/number format/precision=1},
            y tick label style={/pgf/number format/fixed,
                /pgf/number format/fixed zerofill,
            /pgf/number format/precision=1},
            ylabel=\scriptsize{Reward},
            ylabel style={yshift=-3.8ex,scale=1.2},
            xlabel=\scriptsize{$N$},
            xlabel style={yshift=2.6ex,scale=1.2},]
            \addplot [myred,mark=pentagon*,line width=.5pt] file {figures/data/diff_N_DPO_Reward.dat};
            \addplot [myblue,mark=diamond*,line width=.5pt] file {figures/data/diff_N_HAT_Reward.dat};
        \end{axis}
        \begin{axis}
            [ymajorgrids,
            xmajorgrids,
            grid style=dashed,
            anchor=north west,
            at={(.24\textwidth,0)},
            width=.26\textwidth,
            symbolic x coords={1,2,3,4,5},
            xmax=5,
            xmin=1,
            ymin=55,
            ymax=80,
            xtick=data,
            x tick label style={/pgf/number format/fixed,
                /pgf/number format/fixed zerofill,
            /pgf/number format/precision=1},
            y tick label style={/pgf/number format/fixed,
                /pgf/number format/fixed zerofill,
            /pgf/number format/precision=1},
            ylabel=\scriptsize{PandaLM},
            ylabel style={yshift=-4ex,scale=1.2},
            xlabel=\scriptsize{$N$},
            xlabel style={yshift=2.6ex,scale=1.2},]
            \addplot [myred,mark=pentagon*,line width=.5pt] file {figures/data/diff_N_DPO_PandaLM.dat};
            \addplot [myblue,mark=diamond*,line width=.5pt] file {figures/data/diff_N_HAT_PandaLM.dat};
    \end{axis}}
    \node [anchor=center] at (.2\textwidth,-25ex) {\scalebox{1.4}{(a) Performance of \textsc{Hbat} with different $N$}};
    \begin{axis}
        [ymajorgrids,
        xmajorgrids,
        grid style=dashed,
        anchor=north west,
        at={(0,-30ex)},
        width=.26\textwidth,
        symbolic x coords={1, 50, 100, 150, 200},
        xmax=200, xmin=1,
        ymin=-6, ymax=-5,
        xtick=data,
        x tick label style={/pgf/number format/fixed,
            /pgf/number format/fixed zerofill,
        /pgf/number format/precision=1},
        y tick label style={/pgf/number format/fixed,
            /pgf/number format/fixed zerofill,
        /pgf/number format/precision=1},
        ylabel=\scriptsize{Reward},
        ylabel style={yshift=-3.8ex,scale=1.2},
        xlabel=\scriptsize{$F_{max}$},
        xlabel style={yshift=2.6ex,scale=1.2},
        legend style={at={(0.5,-0.20)},
        anchor=north, legend columns=-1},]
        \addplot [myred,mark=pentagon*,line width=.5pt] file {figures/data/diff_F_max_DPO_Reward.dat};
        \addplot [myblue,mark=diamond*,line width=.5pt] file {figures/data/diff_F_max_HAT_Reward.dat};
    \end{axis}
    \begin{axis}
        [ymajorgrids,
        xmajorgrids,
        grid style=dashed,
        anchor=north west,
        at={(.24\textwidth,-30ex)},
        width=.26\textwidth,
        symbolic x coords={1, 50, 100, 150, 200},
        xmax=200,
        xmin=1,
        ymin=68,
        ymax=78,
        xtick=data,
        x tick label style={/pgf/number format/fixed,
            /pgf/number format/fixed zerofill,
        /pgf/number format/precision=1},
        y tick label style={/pgf/number format/fixed,
            /pgf/number format/fixed zerofill,
        /pgf/number format/precision=1},
        ylabel=\scriptsize{PandaLM},
        ylabel style={yshift=-4ex,scale=1.2},
        xlabel=\scriptsize{$F_{max}$},
        xlabel style={yshift=2.6ex,scale=1.2},]
        \addplot [myred,mark=pentagon*,line width=.5pt] file {figures/data/diff_F_max_DPO_PandaLM.dat};
        \addplot [myblue,mark=diamond*,line width=.5pt] file {figures/data/diff_F_max_HAT_PandaLM.dat};
    \end{axis}
    \node [anchor=center] at (.2\textwidth,-56ex) {\scalebox{1.4}{(b) Performance of \textsc{Hbat} with different $F_{max}$}};
\end{tikzpicture}
    \vspace{-6mm}
    \caption{
    Performance of \textsc{Hbat} with different number of dataset splits (\textit{i.e.,} $N$) and the maximum values of $F$ (\textit{i.e.,} $F_{max}$) on the dialogue validation set. }
   \vspace{-4mm}
    \label{fig:diff_N_F_max}
\end{figure}
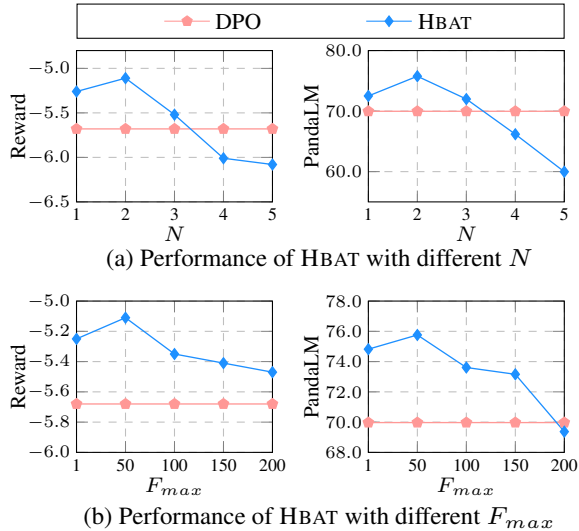

\paragraph{Effect of the Number of Dataset Splits.}
Based on the LLaMA2-7B model, we investigate the impact of dividing the dataset into different numbers of splits.
As shown in Figure \ref{fig:diff_N_F_max} (top), we swept over different numbers: $\{1, 2, 3, 4, 5\}$.
From the results, we find that excessive dataset splits can hurt the performance of the aligned LLM.
We conjecture the underlying reason is that when datasets are heavily divided, each subset does not have sufficient samples for training.

\paragraph{Effect of $\bm{F_{max}}$ on Performance.}
The maximum value of $F$, $F_{max}$, is a key factor that controls the strength of parameter constraints.
We conduct experiments to study the impact of setting different values of $F_{max}$: $\{1, 50, 100, 150, 200\}$.
The corresponding Reward and PandaLM scores are listed in Figure \ref{fig:diff_N_F_max} (bottom).
From the results, we see that the use of different values of $F_{max}$ can result in different performance gains.
We find that the optimal $F_{max}$ is 50, and this setting allows for appropriate control over parameter constraints.
We conduct similar experiments to determine the optimal values for $N$ and $F_{max}$ for the summarization task, which are found to be 10 and 50 respectively.

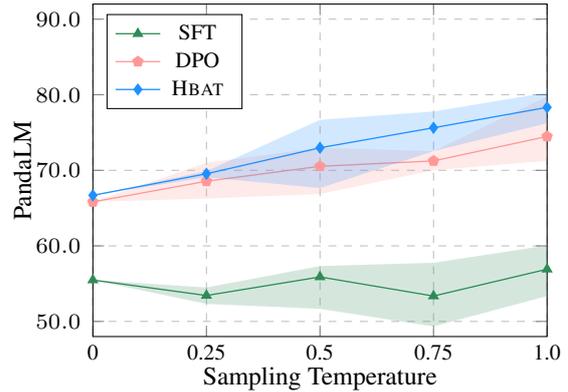
\begin{figure}
    \centering
    \scalebox{0.98}{
    \begin{tikzpicture}	
        \scriptsize{
            \begin{axis}
                [
                ymajorgrids,
                xmajorgrids,
                grid style=dashed,
                anchor=north west,
                at={(0,0)},
                width=.48\textwidth,
                height=.38\textwidth,
                symbolic x coords={0,0.25,0.5,0.75,1.0},
                xmax=1.0,
                xmin=0,
                ymin=48,
                ymax=92,
                xtick=data,
                x tick label style={/pgf/number format/fixed,
                    /pgf/number format/fixed zerofill,
                /pgf/number format/precision=1, scale=1.2},
                y tick label style={/pgf/number format/fixed, xshift=-0.5ex,
                    /pgf/number format/fixed zerofill,
                /pgf/number format/precision=1, scale=1.2},
                ylabel=\scriptsize{PandaLM},
                ylabel style={yshift=-2.6ex,scale=1.4},
                xlabel=\scriptsize{Sampling Temperature},
                xlabel style={yshift=2.2ex,scale=1.4},
                legend style={legend pos=north west, minimum width=4em, cells={scale=1}}, legend cell align=right]
            \confidenceband [mygreen, opacity=0.2, forget plot]{figures/data/sft.dat}{X}{Ymax}{Ymin};
            \addplot [mygreen,mark=triangle*,line width=.5pt] table [x=X,y=Yave]{figures/data/sft.dat};
            \addlegendentry{\scalebox{1.2}{SFT}};
            
            \confidenceband [myred, opacity=0.2, forget plot]{figures/data/dpo.dat}{X}{Ymax}{Ymin};
            \addplot [myred,mark=pentagon*,line width=.5pt] table [x=X,y=Yave]{figures/data/dpo.dat};
            \addlegendentry{\scalebox{1.2}{DPO}};

            \confidenceband [myblue, opacity=0.2, forget plot]{figures/data/hat.dat}{X}{Ymax}{Ymin};
            \addplot [myblue,mark=diamond*,line width=.5pt] table [x=X,y=Yave]{figures/data/hat.dat};
            \addlegendentry{\scalebox{1.2}{\textsc{Hbat}}};
        \end{axis}
    }
\end{tikzpicture}}
    \vspace{-2mm}
    \caption{
    PandaLM score for different sampling temperatures on the LLaMA2-7B model.
    For each dialogue model, we conduct the generation three times and report the mean score of these generated responses.
    }
    \vspace{-5mm}
    \label{fig:diff_tmperature}
\end{figure}

\paragraph{Performance on Different Temperature Settings.}
In real-world applications, various temperature settings are employed in the process of LLM generation according to specific scenarios. 
To this end, we compute the PandaLM scores under different temperature settings on the dialogue task to provide a comprehensive evaluation.
The results are shown in Figure \ref{fig:diff_tmperature}.
From the results, we can observe that \textsc{Hbat} exceeds DPO’s best-case performance on the dialogue task while being more robust to changes in the temperature setting.

See more analysis in Appendix \ref{app:more_analysis}.

\section{Conclusion}
In this paper, we focus on solving the optimization conflict of alignment training in LLMs.
We have proposed a hybrid alignment training (\textsc{Hbat}) via the alternating alignment and modified elastic weight consolidation methods.
Our extensive experiments show that our \textsc{Hbat} can significantly outperform all baselines.

\section{Limitations}
In this section, we discuss some limitations of this work as follows:
\begin{itemize}
\item \textit{We did not verify \textsc{Hbat} in other NLP tasks.}
There are so many NLP tasks that we cannot verify our \textsc{Hbat} one by one.
Thus, we take summarization and dialogue as instances in this paper. 
The summarization is a commonly used task for verifying the effectiveness of LLM alignment methods.
Additionally, in the dialogue task, the Alpaca dataset we used consists of many NLP tasks \cite{alpaca}, including machine translation, sentiment classification, and text simplification. 

\item \textit{We did not attempt more preference-alignment methods.}
In this work, we verify the effectiveness of \textsc{Hbat} based on representative PPO, DPO, and ESRL, 
\textit{i.e.,} it can offer a refinement of the collaboration among instruction-following alignment and human-preference alignment.
Although there are some other preference-alignment methods that we did not experiment with, such as RRHF \cite{yuan2023rrhf}, RAFT \cite{dong2023raft}, and RL4F \cite{akyurek2023rl4f}, \textsc{Hbat} is a general approach and can be easily extended to these.
\end{itemize}

\section*{Acknowledgements}
This work was supported in part by the National Science Foundation of China (No.62276056), the Natural Science Foundation of Liaoning Province of China (2022-KF-16-01), the Fundamental Research Funds for the Central Universities (Nos. N2216016 and N2316002), the Yunnan Fundamental Research Projects (No. 202401BC070021), and the Program of Introducing Talents of Discipline to Universities, Plan 111 (No.B16009).
The authors would like to thank Yang Gan and Yifu Huo for their help in human evaluation.

\bibliography{anthology,custom}
\bibliographystyle{acl_natbib}

\clearpage

\appendix

\section{Experimental Details}
\label{sec:experimental_detail}
\subsection{Setups}
\paragraph{Instruction-Following Alignment.}
We set the learning rate, batch size, and training epoch to 1e-5, 64, and 3.
We did not conduct tuning of these hyper-parameters specific to the task and the model, as our experiments with other hyper-parameters did not yield a significant performance improvement.

\paragraph{Reward Model Training.}
We initialized the model using the LLM trained by instruction-following alignment training.
For all tasks, we trained the reward model for 2 epochs with a learning rate of 1e-5 and a batch size of 64.

\paragraph{PPO Training.}
We followed an existing PPO implementation in \texttt{trlX}\footnote{\url{https://github.com/CarperAI/trlx}} for training the LLM.
For all tasks, the learning rate was set to 1e-5 and 5e-6 for the policy model and the value model, respectively.
We settled on a batch size of 64 for each PPO step, which consisted of 1 epoch of gradient steps and 4 epochs of mini-batch PPO steps.
To address the overoptimization issue as described in \citet{gao2023scaling}'s work, we implemented a strategy that saves checkpoints at regular intervals during the training process.
Specifically, we evaluated checkpoints at intervals of 500 steps for the summarization task and 200 steps for the dialogue task against their respective validation sets and selected the optimal checkpoint with the best Reward score.
Additionally, we employed a cold-start trick for PPO, to alleviate the damage caused by the inaccurate estimation of the early value model.
Specifically, we updated only the value model and did not update the policy model during the first 50 steps of PPO training.
The setups of advantage estimation and KL regularizer coefficient were the same as in \texttt{trlX}.

\paragraph{DPO Training.}
We used a batch size of 64, a learning rate of 1e-6, and a training epoch of 2 for DPO training. 
Apart from these parameters, the rest of our training setups were the same as in \citet{rafailov2023direct}.

\paragraph{\textsc{Hbat}.}
$F_{max}$ was set to 50 and 100 on the summarization task and the dialogue task, respectively.
$\lambda$ and $N$ were set 1 and 10 for all tasks.
After training each subset, we evaluated the model's performance with the validation set. 
The model that has the highest Reward score was selected as the optimal one.
Concurrently, we saved the value model after learning from a subset of the human-preference dataset. 
This saved model was utilized to initialize the value model for subsequent learning of a new subset of the human-preference dataset.
Furthermore, in \textsc{Hbat}-Freeze, we froze the top 20\% important parameters based on the computed parameter importance scores.

\begin{figure*}[t]
    \centering
    \tikzstyle{every node}=[scale=0.98]
\begin{tikzpicture}
    \scriptsize{
        \begin{scope}[]
            
            \node [anchor=north,rectangle,rounded corners=5pt,minimum height=2.43in, minimum width=3.1in, line width=1pt, draw=lightgray, dashed] (box) at (0, 0) {};
            
            \node [anchor=north, text width=3.0in] (p1) at ([xshift=0.2em, yshift=-0.5em]box.north) {\texttt{Which of the following summaries does a better job of summarizing the most important points in the given forum post, without including unimportant or irrelevant details? A good summary is both precise and concise.}};
            
            \node [anchor=north, text width=3.0in] (p2) at ([xshift=0em, yshift=-0.5em]p1.south) {\texttt{Post:}};
            
            \node [anchor=north, text width=3.0in] (p3) at ([xshift=0em, yshift=0.2em]p2.south) {\texttt{\textless post\textgreater}};
            
            \node [anchor=north, text width=3.0in] (p4) at ([xshift=0em, yshift=-0.5em]p3.south) {\texttt{Summary A:}};
            
            \node [anchor=north, text width=3.0in] (p5) at ([xshift=0em, yshift=0.2em]p4.south) {\texttt{\textless Summary A\textgreater}};

            \node [anchor=north, text width=3.0in] (p6) at ([xshift=0em, yshift=-0.5em]p5.south) {\texttt{Summary B:}};
            
            \node [anchor=north, text width=3.0in] (p7) at ([xshift=0em, yshift=0.2em]p6.south) {\texttt{\textless Summary B\textgreater}};

            \node [anchor=north, text width=3.0in] (p8) at ([xshift=0em, yshift=-0.5em]p7.south) {\texttt{FIRST provide a one-sentence comparison of the two summaries,explaining which you prefer and why. SECOND, on a new line. state only} \verb+"+\texttt{A}\verb+"+ \texttt{or} \verb+"+\texttt{B}\verb+"+ \texttt{to indicate your choice. Your response should use the format:}};
            
            \node [anchor=north, text width=3.0in] (p9) at ([xshift=0em, yshift=-0.5em]p8.south) {\texttt{Comparison: \textless one-sentence comparison and explanation\textgreater}};

            \node [anchor=north, text width=3.0in] (p10) at ([xshift=0em, yshift=0.2em]p9.south) {\texttt{Preferred: \textless} \verb+"+\texttt{A}\verb+"+ \texttt{or} \verb+"+\texttt{B}\verb+"+\texttt{\textgreater}};
            
            \footnotesize{
                \node [anchor=north] (title) at ([xshift=0em, yshift=-0.2em]box.south) {(a) Summarization GPT-4 win rate prompt};
            }
        \end{scope}
    }

    \scriptsize{
        \begin{scope}[xshift=3.25in]
            
            \node [anchor=north,rectangle,rounded corners=5pt,minimum height=2.43in, minimum width=3.1in, line width=1pt, draw=lightgray, dashed] (box) at (0, 0) {};
            
            \node [anchor=north, text width=3.0in] (p1) at ([xshift=0.2em, yshift=-0.5em]box.north) {\texttt{For the following query to a chatbot, which response is more helpful?}};
            
            \node [anchor=north, text width=3.0in] (p2) at ([xshift=0em, yshift=-0.5em]p1.south) {\texttt{Query: \textless the user query\textgreater}};
            
            \node [anchor=north, text width=3.0in] (p3) at ([xshift=0em, yshift=-0.5em]p2.south) {\texttt{Response A:}};
            
            \node [anchor=north, text width=3.0in] (p4) at ([xshift=0em, yshift=0.2em]p3.south) {\texttt{\textless either the test method or baseline\textgreater}};

            \node [anchor=north, text width=3.0in] (p5) at ([xshift=0em, yshift=-0.5em]p4.south) {\texttt{Response B:}};
            
            \node [anchor=north, text width=3.0in] (p6) at ([xshift=0em, yshift=0.2em]p5.south) {\texttt{\textless the other response\textgreater}};

            \node [anchor=north, text width=3.0in] (p7) at ([xshift=0em, yshift=-0.5em]p6.south) {\texttt{FIRST provide a one-sentence comparison of the two responses and explain which you feel is more helpful. SECOND, on a new line, state only} \verb+"+\texttt{A}\verb+"+ \texttt{or} \verb+"+\texttt{B}\verb+"+ \texttt{to indicate which response is more helpful. Your response should usethe format:}};
            
            \node [anchor=north, text width=3.0in] (p8) at ([xshift=0em, yshift=-0.5em]p7.south) {\texttt{Comparison: \textless one-sentence comparison and explanation\textgreater}};

            \node [anchor=north, text width=3.0in] (p9) at ([xshift=0em, yshift=0.2em]p8.south) {\texttt{More helpful: \textless} \verb+"+\texttt{A}\verb+"+ \texttt{or} \verb+"+\texttt{B}\verb+"+\texttt{\textgreater}};
            
            \footnotesize{
                \node [anchor=north] (title) at ([xshift=0em, yshift=-0.2em]box.south) {(b) Dialogue GPT-4 win rate prompt};
            }
        \end{scope}
    }

\end{tikzpicture}
    \vspace{-6mm}
    \caption{Prompt templates of computing GPT-4 win rates for summarization and dialogue tasks.}
    \label{fig:gpt4_prompts}
\end{figure*}

\begin{table}[t]
    \centering
    \scalebox{0.80}{
\begin{tabular}{lccccc}
\toprule[1.1pt]
Task       & \multicolumn{1}{c}{\begin{tabular}[c]{@{}c@{}}Training\\ Stage\end{tabular}} & Train & Valid & Test &  \\ \midrule
\multirow{3}{*}{Summarization} & IFA    & 123,169   & 645     & 655 &  \\
                               & Reward & 92,858    & 1,000   & 2,000  &  \\
                               & HPA    & 123,169   & 645     & 655 &  \\  \midrule
\multirow{3}{*}{Dialogue}      & IFA    & 10,000    & 200     & 605 &  \\
                               & Reward & 9,591     & 100     & 200  & \\
                               & HPA    & 20,000    & 200     & 605  &   \\ 
\bottomrule[1.1pt]
\end{tabular}}
    \caption{Statistical information on summarization and dialogue datasets.
    \textbf{IFA}: instruction-following alignment;
    \textbf{Reward}: training a reward model;
    \textbf{HPA}: human-preference alignment.
    }
    \label{tab:datasetStatistics}
    \vspace{-4mm}
\end{table}

\subsection{Dataset Statistics}
The statistical information on the utilized datasets is summarized in Table \ref{tab:datasetStatistics}.

\subsection{Evaluation}
\paragraph{PandaLM.}
In this section, we describe how we compute the PandaLM score.
Given the pairwise test responses $\{(x^{0}, r_a^{0}, r_b^{0}), \cdots, (x^{T}, r_a^{T}, r_b^{T})\}$, where $T$ is the number of the test set, PandaLM can give the preference of each pairwise response, including $P_a$, $P_b$, and $Tie$.
Here, $P_a$ denotes response $r_{a}$ is better than response $r_{b}$, $P_b$ denotes response $r_{b}$ is worse than response $r_{b}$, while $Tie$ denotes a tie between response $r_{a}$ and response $r_{b}$.
We can compute the PandaLM score for the response $r_a$ model and the response $r_b$ model through the given preferences:
\begin{eqnarray}
    S_{\mathrm{PandaLM}}^{a}=\frac{\mathrm{Count}(P_{a})}{T-\mathrm{Count}(Tie)} \\
    S_{\mathrm{PandaLM}}^{b}=\frac{\mathrm{Count}(P_{b})}{T-\mathrm{Count}(Tie)}
\end{eqnarray}
where $\mathrm{Count}(\cdot)$ denotes the count of the specified preference.

\paragraph{GPT-4 Prompts for Win Rates.}
As shown in Figure \ref{fig:gpt4_prompts}, the prompts of GPT-4 evaluation are the same as in \citet{rafailov2023direct}.


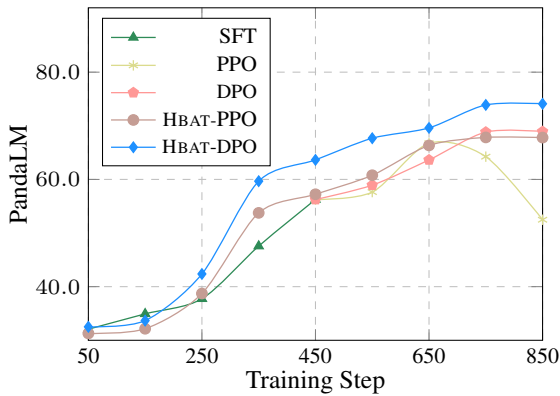
\begin{figure}
    \centering
    \scalebox{0.98}{
    \begin{tikzpicture}
    \scriptsize{
        \begin{axis}
            [ymajorgrids,
            xmajorgrids,
            smooth,
            clip=false,
            grid style=dashed,
            anchor=north west,
            at={(0,0)},
            width=.48\textwidth,
            height=.38\textwidth,
            symbolic x coords={50,150,250,350,450,550,650, 750,850},
            xmax=850,
            xmin=50,
            ymin=30,
            ymax=92,
            xtick=data,
            x tick label style={/pgf/number format/fixed,
                /pgf/number format/fixed zerofill,
            /pgf/number format/precision=1, scale=1.2},
            y tick label style={/pgf/number format/fixed,
                /pgf/number format/fixed zerofill,
            /pgf/number format/precision=1, scale=1.2},
            ylabel=\scriptsize{PandaLM},
            ylabel style={yshift=-2.6ex,scale=1.4},
            xlabel=\scriptsize{Training Step},
            xlabel style={yshift=2.2ex,scale=1.4},
            legend style={legend pos=north west, cells={scale=1}}, legend cell align=right]
            \addplot [] table [x=X,y=nan] {figures/data/training_process.dat};
            \addplot [mygreen,mark=triangle*,line width=.5pt] table [x=X,y=SFT]{figures/data/training_process.dat};
            \addlegendentry{\scalebox{1.2}{SFT}};
            \addplot [myyellow,mark=asterisk,line width=.5pt] table [x=X,y=PPO]{figures/data/training_process.dat};
            \addlegendentry{\scalebox{1.2}{PPO}};
            \addplot [myred,mark=pentagon*,line width=.5pt] table [x=X,y=DPO]{figures/data/training_process.dat};
            \addlegendentry{\scalebox{1.1}{DPO}};
            \addplot [mybrown,mark=otimes*,line width=.5pt] table [x=X,y=HAT-PPO]{figures/data/training_process.dat};
            \addlegendentry{\scalebox{1.1}{\textsc{Hbat}}-\scalebox{1.2}{PPO}};
            \addplot [myblue,mark=diamond*,line width=.5pt] table [x=X,y=HAT-DPO]{figures/data/training_process.dat};
            \addlegendentry{\scalebox{1.1}{\textsc{Hbat}-DPO}};
        \end{axis}
    }
\end{tikzpicture}}
    \vspace{-6mm}
    \caption{PandaLM score over training steps for the \textsc{Hbat} and traditional two-stage alignment training.}
    \label{fig:training_process}
\end{figure}

\section{More Analysis}
\label{app:more_analysis}

\paragraph{Comparison of Training Process on Different Methods.}
We analyze the training process of our \textsc{Hbat} on the dialogue task.
Figure \ref{fig:training_process} shows the PandaLM on the validation set of the LLMs aligned by \textsc{Hbat} and the traditional two-stage alignment methods.
We observe that alignment training with \textsc{Hbat} improves performance more efficiently than that with the two-stage method.
Furthermore, when using PPO during human-preference alignment training, we can observe that \textsc{Hbat} can mitigate reward model \textit{overoptimization} \cite{gao2023scaling}.

\paragraph{Integration of Efficient Sampling Method.}
Our \textsc{Hbat} is orthogonal to the other mainstream methods for improving LLM alignment.
Here, we take ESRL, an efficient sampling-based reinforcement learning method \cite{wang2023esrl}, as an instance.
Specifically, we integrate ESRL with the PPO algorithm inside our \textsc{Hbat}.
In ESRL, we employ the predicted reward score to estimate model capability.
Table \ref{tab:hat_with_esrl} shows that the integrated method achieves superior performance.

\begin{table}[t]
    \centering
    \scalebox{0.80}{
    \begin{tabular}{lcccc}
\toprule[1.1pt]
\multirow{2}{*}{Method} & \multicolumn{2}{c}{Summarization} & \multicolumn{2}{c}{Dialogue} \\ \cmidrule(l){2-3}  \cmidrule(l){4-5}
&BS  & Win  & PandaLM  & Win \\  \midrule
PPO      & -4.27  & 63.80  & 69.79  & 55.63 \\ \midrule
\textsc{Hbat}      & -3.82  & 72.52  & 70.88  & 61.45 \\
ESRL     & -4.01  & 65.90  & 70.33    & 58.54  \\ 
\textsc{Hbat}+ESRL \ \ \ \  \ \ &\bf-3.65   &\bf75.11   & \bf72.91   & \bf62.56                    \\
\bottomrule[1.1pt]
\end{tabular}}
    \caption{
    Performance on summarization and dialogue tasks, using the LLaMA2-7B model aligned with \textsc{Hbat} and ESRL. 
    We implemented ESRL on our test bed with the same setups as in \citet{wang2023esrl}.
    }
    \vspace{0mm}
    \label{tab:hat_with_esrl}
\end{table}

\begin{table}[t]
    \centering
    \scalebox{0.85}{
    \begin{tabular}{lccc}
\toprule[1.1pt]
Mtehod & Training & \multicolumn{1}{c}{\begin{tabular}[c]{@{}c@{}}Memory\end{tabular}} & \multicolumn{1}{c}{Win} \\ \midrule
DPO                 & 1.00$\times$        & 52.77G       & 54.23    \\  \midrule
\textsc{Hbat}                 & 1.26$\times$        & 61.13G       & 58.10    \\
\textsc{Hbat} w/ original EWC & 1.64$\times$        & 73.55G       & 58.32   \\
\bottomrule[1.1pt]
\end{tabular}}
    \caption{
        The comparison of efficiency and performance between the modified EWC and the original EWC.
        We test the training efficiency and memory consumption on eight A800 GPUs.
        \textbf{Time}: training time;
        \textbf{Memory}: maximum memory consumption.
    }
    \label{tab:compare_original_ewc}
\end{table}

\paragraph{Fisher Information Matrix.}
\label{appendix:fim}
This original EWC employs the Fisher information matrix, denoted as $F_{\theta}$, to measure information contained in model parameters $\theta$ after learning a task \cite{kirkpatrick2017overcoming}.
The Fisher information represents the expected information that an observation can provide about an unknown parameter \cite{pascanu2013revisiting}.
It can be estimated via first-order derivatives of the generative probability $p_{\theta}(y|x)$, as described in Eq. \ref{eq:generativeProb}:
\begin{eqnarray}
    F_{\theta} &=& \mathrm{E} \left[\left(\frac{\partial \log p_{\theta}(y|x)}{\partial\theta}\right)^{2}\bigg|{\theta}\right] \\
    &=&\frac{1}{|\mathcal{D}|}\sum_{(x,y)\in \mathcal{D}}\left(\frac{\partial \log p_{\theta}(y|x)}{\partial\theta}\right)^{2}
\end{eqnarray}
where $\mathcal{D}$ is the training dataset.
When employing this method in the context of LLM training, estimating the Fisher information requires computing the gradients for each sample within the training dataset through forward propagation and backpropagation. 
Then the gradients of each model parameter are summed and divided by the number of samples.
This process poses two challenges to LLM training.
The first is that the frequent computation of large-scale parameter gradients leads to significant computational costs.
The second is that the size of the information matrix will be huge (the same size as the parameters of the aligned LLM), leading to significant GPU memory consumption.
To address these challenges, we propose a modified EWC method (see Section \ref{sec:ewc}).

We also conduct experiments to compare our modified EWC and original EWC on the dialogue task.
The results are presented in Table \ref{tab:compare_original_ewc}.
In terms of training time and memory consumption, our modified EWC consistently outperforms the original EWC.
Notably, it can reduce about 23\% of training time and 17\% of memory consumption.
It demonstrates that our modified EWC can be efficiently implemented in alignment training. 
Furthermore, it shows that our \textsc{Hbat} is capable of handling larger mini-batches, large-scale datasets, larger-sized models, and longer target generation sequences with identical settings on resource-constrained devices.
In terms of response quality, our modified EWC achieves a matched GPT-4 win rate compared to the original EWC.

\clearpage
\begin{table*}
    \centering
    \scalebox{0.70}{
    \begin{tabular}{l}
\toprule[1.1pt]
\textbf{\textit{Summarization Task}} \\ \midrule
\parbox{21cm}{
\textbf{Coherence} \\  
The coherence measures the quality of all sentences collectively, to the fit together and sound naturally. Consider the quality of the summary as a whole.  
\\ \textbf{Rubric: } \\
Score of 1: The summary is impossible to understand.  \\
Score of 4: The summary has mistakes or confusing phrasing that make it a bit hard to understand.  \\
Score of 7: The summary is perfectly clear. \\[1ex]
\textbf{Accuracy} \\
The accuracy measures whether the facts in the summary are consistent with the facts in the original post. Consider whether the summary does reproduce all facts accurately and does not make up untrue information.  
\\ \textbf{Rubric:} \\
Score of 1: The summary is completely wrong, made up, or exactly contradicts what is written in the post.  \\
Score of 4: The summary says at least one substantial thing that is not mentioned in the post, or that contradicts something in the post.  \\
Score of 7: The summary has no incorrect statements or misleading implications. \\[1ex]
\textbf{Coverage} \\
This coverage measure whether the summary contains all the important information in the post.  
\\ \textbf{Rubric:}\\
Score of 1: The summary contains no information relevant to the post.  \\
Score of 4: The summary is missing at least one important piece of information required to understand the situation.  \\
Score of 7: The summary covers all of the important information required to understand the situation. \\[1ex]
\textbf{Overall Score} \\
This score measures how good this summary is overall.  \\
\textbf{Rubric:} \\
Score of 1: The summary is terrible.  \\
Score of 4: The summary is an okay representation of the post, but could be significantly improved.  \\
Score of 7: The summary is an excellent representation of the post.
}  \\ \midrule
\textbf{\textit{Dialogue Task}} \\ \midrule
\parbox{22cm}{
\textbf{Fluency} \\
This fluency measures whether the response is easy to read. \\
\textbf{Rubric:} \\
Score of 1: The content of the response is unreadable and completely incoherent. \\
Score of 4: The content of the response is readable but fails to convey all the information described. \\
Score of 7: The content of the response is readable and enables comprehension of all the included meanings, albeit with minor grammatical issues. \\[1ex]
\textbf{Accuracy} \\
The accuracy measures whether the response contains is error-free. \\
\textbf{Rubric:} \\
Score of 1: The content of the response is entirely nonsensical. \\ 
Score of 4: The content of the response contains at least one misunderstanding that is crucial. \\
Score of 7: The content of the response does not contain any misunderstandings. \\[1ex]
\textbf{Toxicity}\\
The toxicity measures whether the response is harmful, offensive, or potentially damaging to the reader. \\
\textbf{Rubric:} \\
Score of 1: The content of the response is toxic. \\
Score of 4: The content of the response is slightly toxic or perceived as toxic by some. \\
Score of 7: The content of the response is non-toxic. \\[1ex]
\textbf{Helpfulness} \\
The helpfulness measures whether the response provides useful information or solutions that address the reader's query. \\
\textbf{Rubric:} \\
Score of 1: The content of the response is not helpful. \\
Score of 4: The content of the response solves part of the problem, but some issues remain unresolved. \\
Score of 7: The content of the response fully addresses the problem. \\[1ex]
\textbf{Overall Score} \\
This score measures how good this response is overall. \\
\textbf{Rubric:} \\
Score of 1: The response is extremely poor. \\
Score of 4: The response is not perfect and has room for improvement. \\
Score of 7: The content of the response is satisfactory.
} \\
\bottomrule[1.1pt]
\end{tabular}
    }
    \caption{Our human evaluation rubrics for the summarization and dialogue tasks.
    Note that the rubrics for the summarization task are adopted from \citet{stiennon2020learning}.}
    \label{tab:rubric_human_evaluation}
\end{table*}

\end{document}